\documentclass[%
 aip,
 amsmath,amssymb,
 reprint,%
]{revtex4-1}

\usepackage{graphicx}
\usepackage{dcolumn}
\usepackage{bm}

\usepackage[utf8]{inputenc}
\usepackage[T1]{fontenc}
\usepackage{mathptmx}
\usepackage{etoolbox}
\usepackage{xcolor}
\newcommand{\jz}[1]{\textcolor{black}{#1}}

\makeatletter
\def\@email#1#2{%
 \endgroup
 \patchcmd{\titleblock@produce}
  {\frontmatter@RRAPformat}
  {\frontmatter@RRAPformat{\produce@RRAP{*#1\href{mailto:#2}{#2}}}\frontmatter@RRAPformat}
  {}{}
}%
\makeatother
\begin{document}

\preprint{AIP/123-QED}

\title{Physics-informed Active Polarimetric 3D Imaging for Specular Surfaces}
\author{Jiazhang Wang}
\affiliation{Wyant College of Optical Sciences, University of Arizona, Tucson, AZ 85721}
\email{jiazhangwang@arizona.edu, fwillomitzer@arizona.edu}
\author{Hyelim Yang}%
\affiliation{Wyant College of Optical Sciences, University of Arizona, Tucson, AZ 85721}

\author{Tianyi Wang}
\affiliation{Wyant College of Optical Sciences, University of Arizona, Tucson, AZ 85721}

\author{Florian Willomitzer}
\affiliation{Wyant College of Optical Sciences, University of Arizona, Tucson, AZ 85721}

\date{\today}

\begin{abstract}
3D imaging of specular surfaces remains challenging in real-world scenarios, such as in-line inspection or hand-held scanning,  requiring fast and accurate measurement of complex geometries. Optical metrology techniques such as deflectometry achieve high accuracy but typically rely on multi-shot acquisition, making them unsuitable for dynamic environments. Fourier-based single-shot approaches alleviate this constraint, yet their performance deteriorates when measuring surfaces with high spatial frequency structure or large curvature. Alternatively, polarimetric 3D imaging in computer vision operates in a single-shot fashion and exhibits robustness to geometric complexity. However, its accuracy is fundamentally limited by the orthographic imaging assumption.

In this paper, we propose a physics-informed deep learning framework for single-shot 3D imaging of complex specular surfaces. Polarization cues provide orientation priors that assist in interpreting geometric information encoded by structured illumination. These complementary cues are processed through a dual-encoder architecture with mutual feature modulation, allowing the network to resolve their nonlinear coupling and directly infer surface normals. The proposed method achieves accurate and robust normal estimation in single-shot with fast inference, enabling practical 3D imaging of complex specular surfaces.
\end{abstract}

\maketitle

\begin{figure*}[t]
\centering
\includegraphics[width=\linewidth]{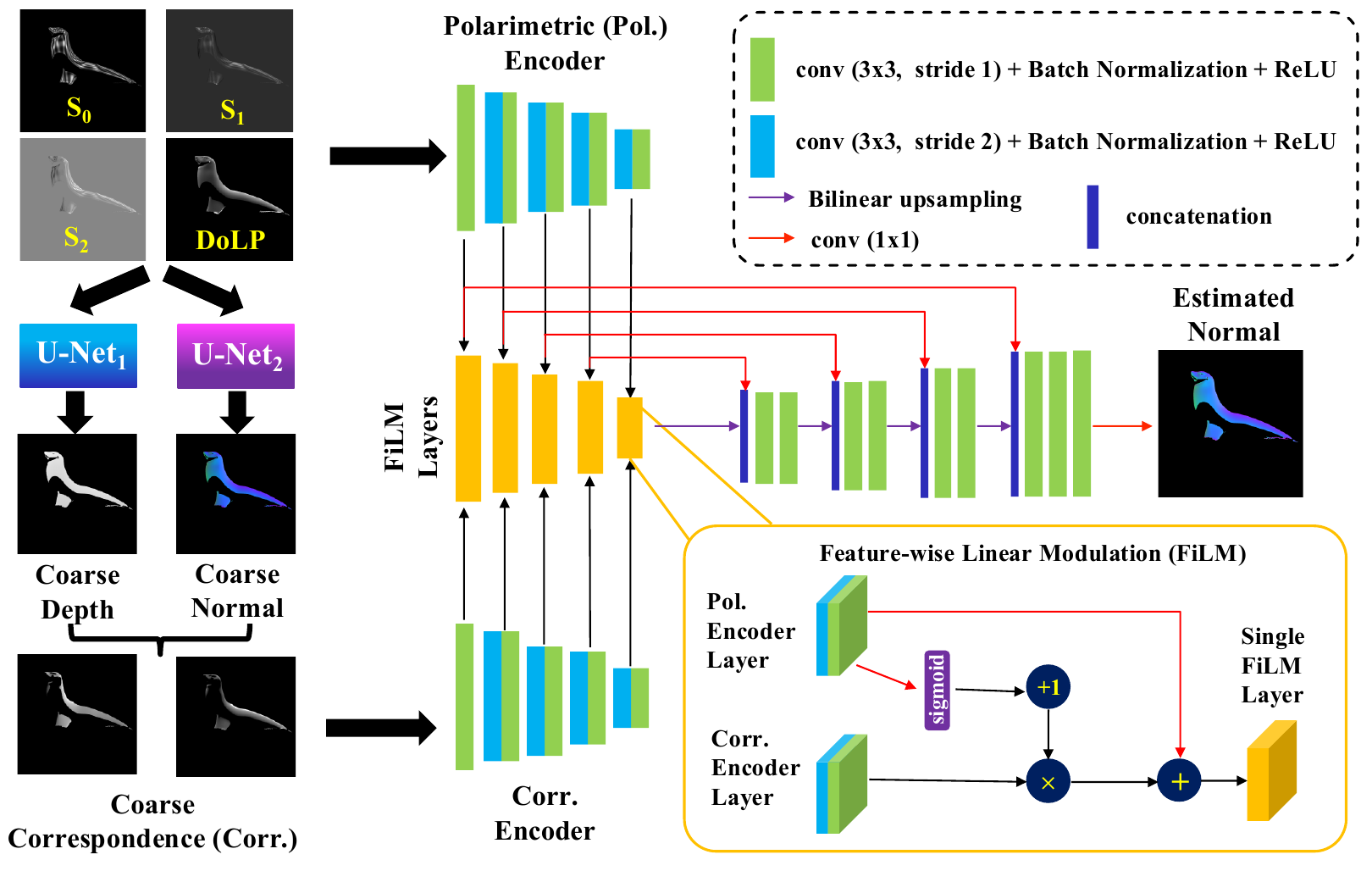}
\caption{\textbf{Overview of the proposed physics-informed learning framework.} 
Polarimetric inputs, including Stokes parameters and DoLP, are first processed by U-Nets to obtain coarse depth and normal estimates. Then the coarse correspondence map is analytically calculated.
These physics priors are processed separately through two encoder branches to extract modality-specific features
Feature-wise Linear Modulation layers are employed to adaptively fuse polarimetric cues and geometric correspondence features, enabling robust normal estimation. 
}
\label{model}
\end{figure*}

\section{Introduction}

Accurate and robust 3D imaging of specular surfaces plays a critical role in a wide range of applications, including industrial inspection~\cite{burke2023deflectometry}, cultural heritage preservation~\cite{willomitzer2020hand}, and robotic perception~\cite{li2025robotic}. Nevertheless, achieving fast and reliable reconstruction in real-world scenarios remains challenging. First, the motion-robust measurements are required in applications such as in-line inspection on moving conveyor belts or hand-held 3D scanning systems~\cite{willomitzer2020hand}. In addition, the target objects often exhibit complex geometries, especially large curvature and high spatial frequency. Both of them pose significant challenges to existing methods. 

In optical metrology, deflectometry~\cite{knauer2004phase,wang2025accurate,dashpute2023event} is a representative technique that provides high-accuracy measurement of specular surfaces. Its performance depends on the precise camera-screen correspondence, which is typically established through multi-shot acquisition of sequential structured light patterns, such as phase-shifting methods~\cite{zuo2018phase,huang2018review}. This sequential strategy, however, is incompatible with high-speed or dynamic measurement scenarios. Single-shot deflectometry approaches have been demonstrated~\cite{liang2020using,wang2025accurate}, for example, by displaying a cross-sinusoidal pattern and retrieving the correspondence via Fourier-based analysis. However, the complex surface geometries targeted in this work strongly deviate from the low frequency object (e.g., ``spherical lens shape'') that is typically measured with deflectometry. The high spatial frequency or large curvature of these surfaces induce large frequency  variation of the reflected pattern. As a result, Fourier-based methods suffer from bandwidth limitations, leading to performance degradation. Although windowed Fourier transform~\cite{Kemao:04} or wavelet transform-based methods~\cite{liang2020using,wang2025accurate} has been explored to alleviate this issue, such approaches often lack robustness in the presence of noise and introduce substantial computational overhead which prevents potential real-time evaluation.  Even if the phase can be successfully retrieved, phase unwrapping is required to recover continuous camera-screen correspondence. This process typically relies on additional patterns, compromising the single-shot nature of the system, or requires strong surface priors, further limiting the practicality and robustness of standard deflectometry.  Moreover, the inherent depth–normal ambiguity~\cite{wang2025accurate,wang20253d} in single-camera deflectometry further restricts its applicability to full-field 3D measurements. Although stereo deflectometry~\cite{knauer2004phase,wang2025accurate} resolves this ambiguity, it significantly reduces the effective measurement region.

As an alternative, computer vision techniques offer greater flexibility and robustness in practical scenarios and are generally less sensitive to geometric complexity. Among them, polarimetric 3D imaging~\cite{rahmann2001reconstruction,miyazaki2003polarization,atkinson2006recovery,kadambi2015polarized,smith2018height,ba2020deep,lei2022shape,muglikar2023event,peng2026joint} enables single-shot measurement of specular surfaces in a passive manner by exploiting polarization cues. However, besides the long-standing normal ambiguity issue~\cite{atkinson2006recovery,kadambi2015polarized}, its measurement accuracy is fundamentally constrained by the oversimplified orthographic imaging assumption, which treats reflected rays as perpendicular to the image plane. This simplification conflicts with the physics of perspective imaging present in most systems. As a result, significant surface normal errors can arise, typically exceeding 5° and in some cases reaching several tens of degrees. 
\jz{Recently, data-driven deep learning approaches \cite{ba2020deep,lei2022shape,muglikar2023event,peng2026joint} have been introduced to improve the performance of conventional shape-from-polarization (SfP). By learning from large datasets and polarimetric priors, these methods aim to mitigate several long-standing challenges in polarimetric 3D imaging, such as normal ambiguity, varying material properties, and image noise. However, the fundamental principle behind most learning-based SfP approaches still relies on the simplified orthographic imaging assumption inherited from classical SfP models, which fundamentally limits their achievable accuracy. Moreover, these \jz{DL} methods are primarily designed for diffuse or mixed-reflectance surfaces, and are not specifically tailored for specular objects.}
These limitations hinder the adoption of polarimetric imaging in applications of measuring specular surfaces demanding high accuracy, such as robotic manufacturing and medical imaging. 

\jz{The} previous work~\cite{wang20253d} introduced a multi-modal 3D imaging framework that integrates polarimetric and geometric cues, enabling surface reconstruction without reliance on the orthographic imaging assumption. Despite these advantages, the method remains limited in practical applications. First, it still encounters challenges in establishing reliable camera-screen correspondence under single-shot acquisition when measuring complex geometries with high spatial frequencies or large curvature. Second, as a purely analytical formulation that tightly couples polarimetric and geometric measurements, the reconstruction follows a deterministic computation pipeline. Noise or estimation errors in either modality directly propagate through this pipeline and may be amplified, leading to degraded performance.

In this contribution, we address these issues by leveraging a physics-informed deep neural network. In the first stage, instead of explicitly computing the camera-screen correspondence, we input the polarization priors to the network to estimate the correspondence map. In the second stage, physics priors (polarization and geometric cues) are processed by two separate encoder branches to extract complementary features. These features are subsequently fused in a shared decoder, where cross-modal feature modulation is introduced to mitigate error propagation from individual modalities. The proposed method exhibits strong robustness to unseen objects while maintaining high accuracy. The mean angular error of the estimated normals is $0.79^\circ$, significantly outperforming conventional polarimetric 3D imaging approaches. Its single-shot capability and fast inference improve the practicality and efficiency compared to conventional optical metrology systems.

\section{Method}

The proposed method jointly exploits polarimetric and geometric information to estimate the surface normals of specular objects. Polarimetric cues encode surface orientation by providing the reflection angle between the surface normal and the viewing direction. Geometric information, obtained from the screen's structured illumination, establishes the correspondence between the light source emission point (screen pixel) and the camera pixel observing the reflected ray. In ideal conditions, these complementary modalities uniquely determine the surface shape. A detailed description of how the  multimodal information is acquired is presented in our previous work~\cite{wang20253d}, which used an analytical formulation for evaluation.

In this work, we adopt the same system configuration consisting of an unpolarized display screen and a polarization camera capable of capturing four images at different polarization angles in a single-shot. The first stage of the proposed method aims to mitigate the difficulty of establishing camera–screen correspondence from a single-shot pattern (e.g., a cross-sinusoidal pattern). Instead of explicitly analyzing the deformed fringe pattern in the frequency domain, we leverage polarimetric cues.
Given four polarization images $I_0, I_{45}, I_{90}, I_{135}$ captured at polarizer angles $0^\circ, 45^\circ, 90^\circ,$ and $135^\circ$, we compute the Stokes parameters and the degree of linear polarization (DoLP) as follows:
\begin{align}
S_0 &= \frac{1}{2}(I_0 + I_{45} + I_{90} + I_{135}), \quad
S_1 = I_0 - I_{90}, \\
S_2 &= I_{45} - I_{135}, \quad
\mathrm{DoLP} = \frac{\sqrt{S_1^2 + S_2^2}}{S_0}.
\end{align}
Beyond encoding surface orientation, the Stokes vectors also preserve the pattern deformation. This deformation, induced by specular reflection, implicitly contains geometric information about surface shape. Therefore, the polarimetric representation not only provides surface orientation priors but also carries active-illumination-induced geometric cues.

These polarimetric physics priors are fed into two separate U-Net models to predict coarse surface depth and normals (see Fig.~\ref{model}). According to the law of specular reflection, and given the predicted depth and normal together with calibrated camera intrinsics, the viewing direction and corresponding incident ray can be determined. With the calibrated screen pose, the intersection between the incident ray and the screen plane is computed to recover the corresponding screen emission pixel.

The second stage aims to reduce error amplification caused by noise and uncertainty, which may propagate through the analytical reconstruction pipeline. To more effectively exploit the complementary physics information (polarimetric and geometric cues) while mitigating error propagation, we feed the data into two independent encoder branches (polarimetric encoder and correspondence encoder) to extract modality-specific features (see Fig.\ref{model}).

The geometric cues, represented by the coarse correspondence map, may be unreliable in regions with high spatial frequency or large curvature. To address this issue, we introduce a feature-wise linear modulation (FiLM) layer~\cite{perez2018film}, where polarization features are used to modulate the geometric features. This mechanism enables the network to adaptively weight geometric information according to the local polarization state, thereby suppressing unreliable geometric estimates.
The modulated geometric features are then fused with polarization features in a shared decoder to predict the final surface normal map. The network is trained using a masked mean angular error loss and optimized with Adam optimizer, with an initial learning rate of $1\times10^{-4}$ scheduled by cosine annealing.

Data quality plays a critical role in the performance of learning-based 3D imaging methods.  \jz{However, publicly available datasets for specular objects are limited.This challenge becomes even more significant in our case, as the input to our framework is fundamentally different from conventional methods. Rather than relying on raw intensity images, our approach requires physics-informed inputs, including reflected structured light patterns and polarization cues. As a result, existing datasets are not directly applicable, making it necessary to construct a dedicated dataset tailored to our problem.}However, acquiring reliable ground-truth data for challenging real world specular surfaces remains difficult for several reasons. First, most off-the-shelf specular objects do not provide ground-truth surface normals. Second, ultra-high-precision measurement systems, such as stereo deflectometry or interferometry, are difficult to deploy in large-scale data collection due to their complexity and limited throughput. Third, many learning-based polarimetric methods rely on commercial 3D scanners to obtain surface depth and subsequently compute normals via numerical differentiation. However, most commercial 3D scanners (e.g., stereo vision or time-of-flight systems) are primarily designed for diffuse reflection–dominated surfaces and typically perform poorly on specular objects. In addition, their accuracy is often insufficient for high-precision tasks and may not exceed the performance level targeted by our method. Even when depth can be recovered, numerically differentiating depth maps often introduces noise, making the resulting normals unreliable as ground truth.

\begin{figure}[t]
\centering
\includegraphics[width=0.8\linewidth]{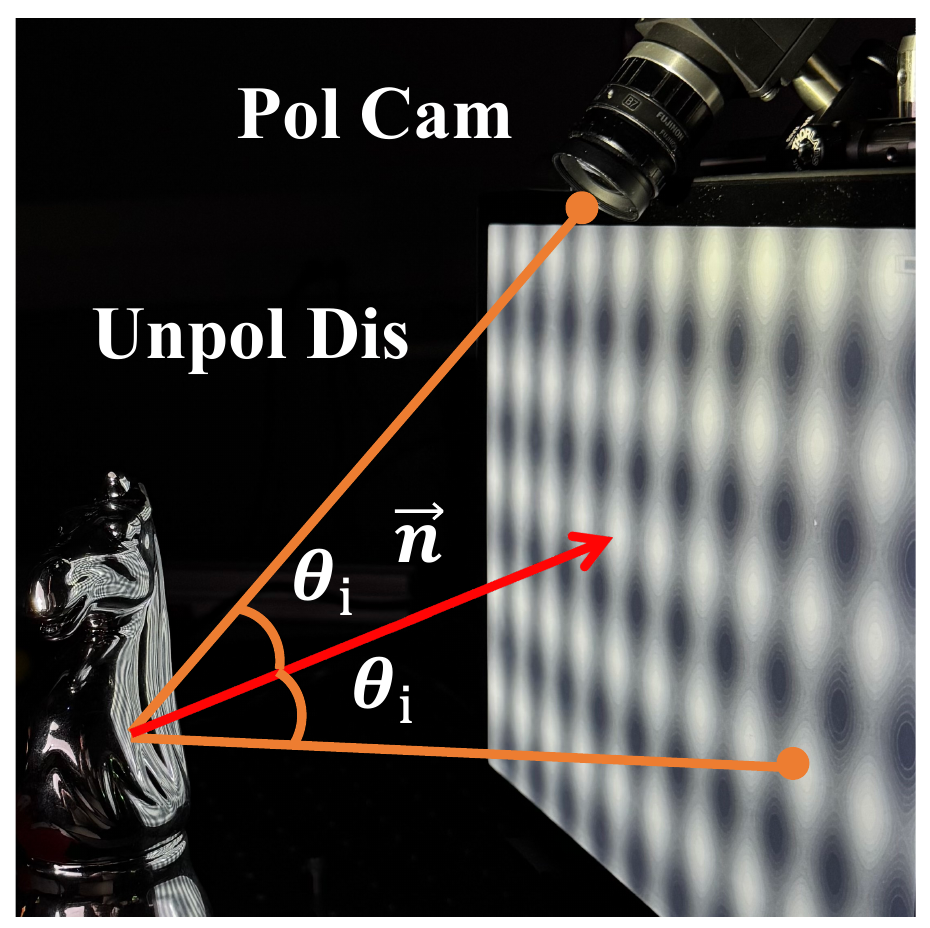}
\caption{\textbf{Real experimental prototype.}
The prototype consists of a polarization camera and an unpolarized display. The polarization cues encode the surface orientation information. Decoding the reflected deformed stucture light pattern reveals the surface geometrical cues. }
\label{setup}
\end{figure}

\begin{figure*}[t]
\centering
\includegraphics[width=\linewidth]{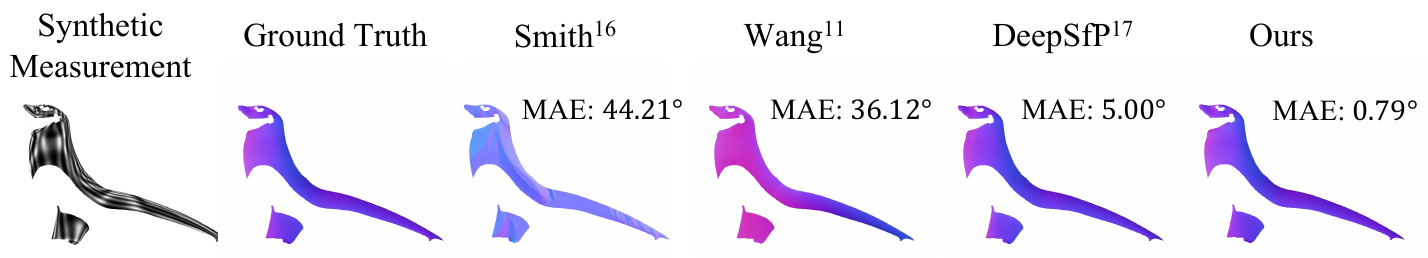}
\caption{\textbf{Normal comparison with different polarimetric 3D imaging methods.} From left to right: measurement image, ground truth, the classical single-view physical SfP method of Smith~\cite{smith2018height}, the previous multi-modal physics-based method Wang~\cite{wang20253d}, the learning-based method DeepSfP~\cite{ba2020deep}, and the proposed method. The Mean Angular Error (MAE) for each method is reported. Physics-based methods suffer from large reconstruction errors due to inherent modeling issues, while the learning-based DeepSfP improves the performance but is still limited by the orthographic imaging assumption. In contrast, the proposed method leverages complementary physical priors to achieve significantly improved accuracy, reducing the MAE to $0.79^\circ$.
}
\label{seal}
\end{figure*}

To address these challenges, we generate training data using a physics-based rendering engine, Mitsuba \cite{Mitsuba3}. We construct a digital twin of our experimental setup (see Fig.~\ref{setup}(a)) in a virtual environment, replicating the camera intrinsics, screen pose, and illumination patterns. This enables the synthesis of physically consistent polarization images along with accurate ground-truth surface normals.
The dataset consists of 38 distinct 3D objects collected online. Each object is rendered under varying poses and viewpoints, resulting in 605 unique samples. All images are generated at a resolution of $1024 \times 1024$ for four polarization angles consistent with our real polarization camera setup. To better approximate real measurements, random noise is added to each rendered image, with the signal-to-noise ratio controlled between $40dB$ and $50dB$.

\section{Experiments and Discussion}
Fig.~\ref{setup} shows our real setup consisting of a polarization camera (model: FLIR BFS-US-51S5PC-C) and an unpolarized display (model: BOOX Tablet Tab). All experiments are conducted on a workstation equipped with an Intel Core i9-14900F CPU, 32GB of memory, and an NVIDIA GeForce RTX 5070 Ti GPU.

\begin{figure}[b]
\centering
\includegraphics[width=\linewidth]{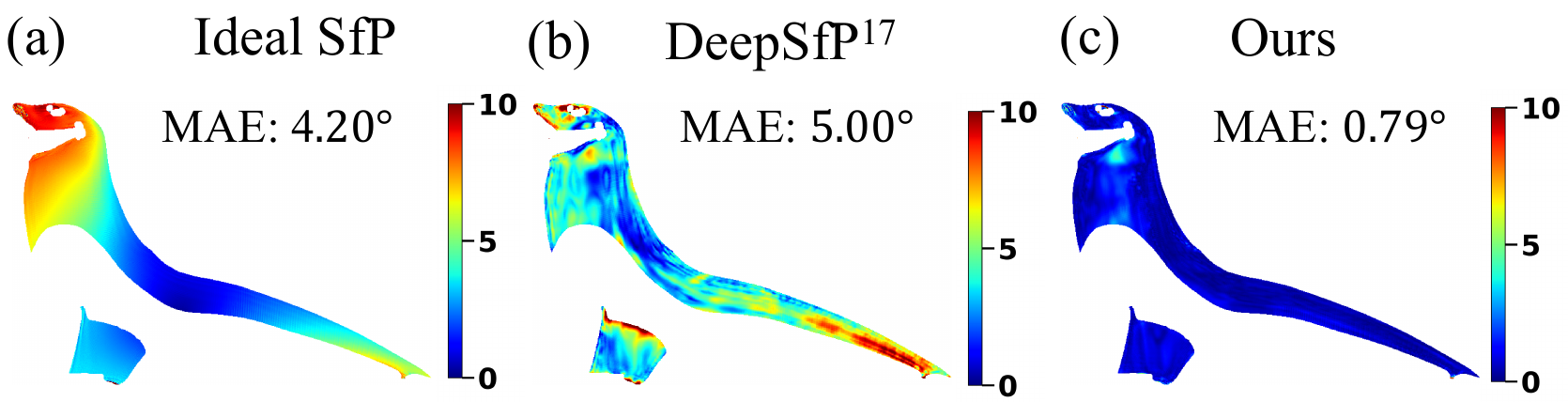}
\caption{\textbf{Error maps of surface normal estimation.} (a) Ideal SfP where the normal ambiguity is manually resolved, leaving the orthographic imaging assumption as the primary error source. The error increases toward the image periphery. (b) DeepSfP~\cite{ba2020deep} exhibits a similar spatial error distribution, indicating that learning-based SfP methods remain constrained by the same physical limitation. (c) The proposed method significantly reduces the reconstruction error across the entire surface by incorporating complementary physical priors, achieving a Mean Angular Error (MAE) of $0.79^\circ$.
}
\label{sfp}
\end{figure}

To evaluate the performance of the proposed network, we \jz{first} test it on a set of unseen objects that is not included in the training dataset and \jz{compare the reconstructions against several state-of-the-art techniques}. \jz{Fig.~\ref{seal} demonstrates the reconstructed surface normal using various polarimetric 3D imaging methods. Smith~\cite{smith2018height}, a classical single-view physics-based approach, is frequently adopted as a baseline in polarimetric 3D imaging studies. For our test object, it yields a Mean Angular Error (MAE) of $44.21^\circ$. The previous physics-based method Wang \cite{wang20253d} can achieve high accuracy by jointly exploiting polarimetric cues and geometric cues under multi-shot acquisition. However, in a single-shot scenario it becomes difficult to reliably decode the highly deformed reflected structured-light patterns at the high slope parts of the object to recover accurate geometric cues, resulting in an MAE of $37.40^\circ$. DeepSfP \cite{ba2020deep} represents the first learning-based polarimetric 3D imaging method. It takes candidate surface normals derived from conventional SfP under diffuse and specular assumptions as physical priors. Since our task focuses on specular objects, we only input the specular prior for comparison, which in fact simplifies the problem. DeepSfP achieves an MAE of $5.00^\circ$. In contrast, our method significantly outperforms all existing approaches, achieving an MAE of only $0.79^\circ$. 
We further compare DeepSfP with our method in Fig.~\ref{sfp} by visualizing the error maps. For reference, we also evaluate an ideal SfP scenario in which the normal ambiguity is manually resolved, leaving the orthographic imaging assumption as the only source of error. As shown, the error of ideal SfP increases as the measurement points move away from the image center. The error of DeepSfP exhibits the same trend of increasing error toward peripheral regions. This observation indicates that learning-based SfP methods are still fundamentally constrained by the inherent physical limitations of SfP. In contrast, our method leverages complementary physical priors to overcome this limitation, enabling accurate surface normal reconstruction beyond the constraints of the conventional SfP model. Moreover, compared with Wang\cite{wang20253d}, our method demonstrates high robustness in single-shot measurement of complex geometry. }  

\begin{figure*}[t]
\centering
\includegraphics[width=\linewidth]{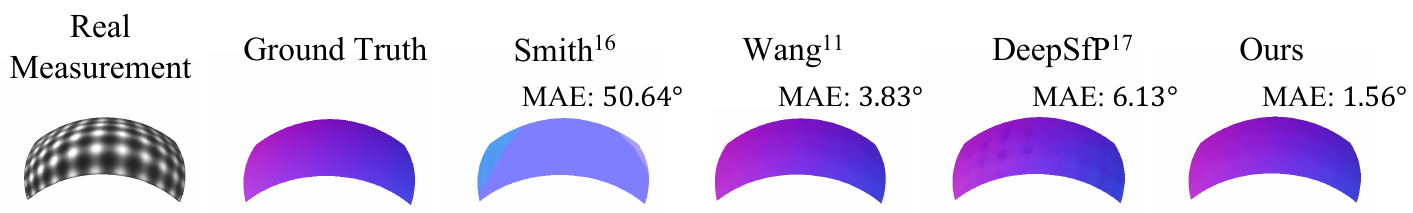}
\caption{\textbf{Quantitative evaluation on real measurement.} We perform quantitative evaluation using a precisely manufactured ball bearing with a known radius of 1 inch. Our proposed method achieves an MAE of $1.56^\circ$, outperforming previous methods. It is worth noting that while the previous multi-modal physics-based method by Wang~\cite{wang20253d} also demonstrated reconstruction of a ball bearing, in our experiment, the object has smaller size, which leads to larger angular coverage and more severe spatial pattern deformation, making single-shot correspondence retrieval significantly more challenging and leading to degraded performance of the physics-based method.
}
\label{real}
\end{figure*}

\begin{figure}[h]
\centering
\includegraphics[width=0.8\linewidth]{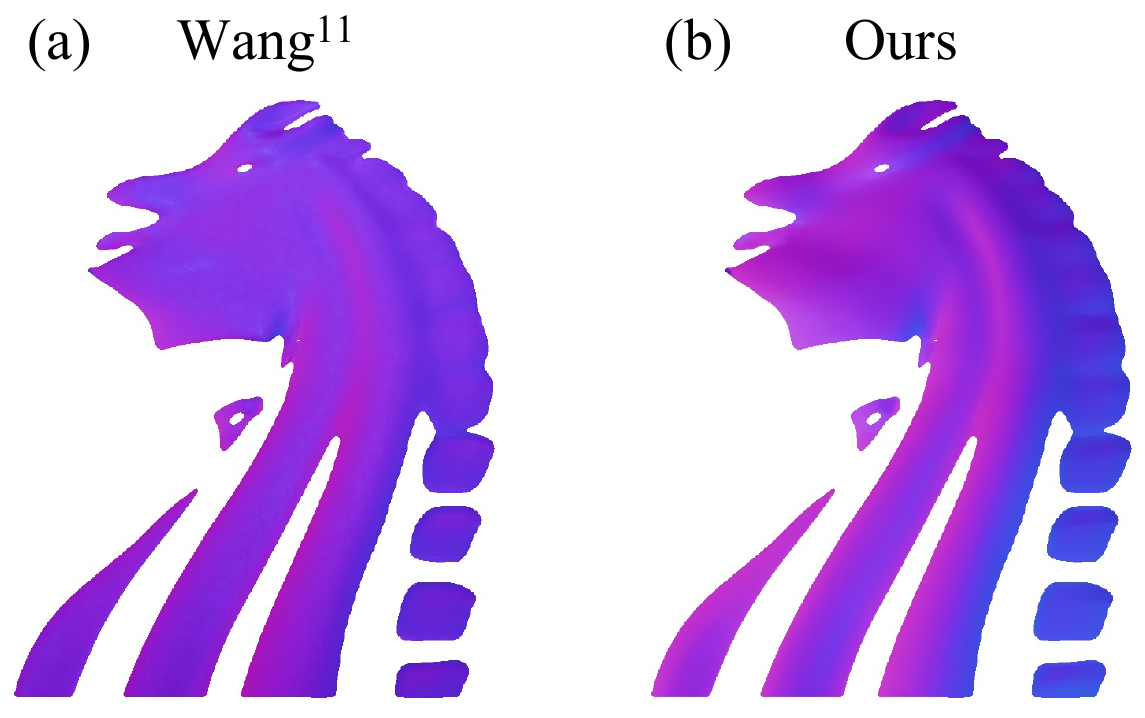}
\caption{\textbf{Qualitative evaluation on complex real object.} 
(a) Surface normal estimated using the previous physics-based method~\cite{wang20253d} with \textbf{multi-shot} sequential captures. (b) Surface normal estimated by the proposed method with \textbf{single-shot} capture. The proposed method produces a more consistent normal field, especially in the cheek region of the horse.
}
\label{horse}
\end{figure}

\jz{To further validate the performance of the proposed network on real-world data, we input the images captured with our real (see fig.~\ref{setup}). Because it's challenging to obtain the reliable ground truth of specular surfaces, We first conduct a quantitative evaluation on a precisely manufactured ball bearing with known radius of 1 inch. Comparing with the similar experiment conducted in previous physics-based method\cite{wang20253d}, we use a smaller ball bearing, which leads to stronger spatial pattern deformation, where single-shot correspondence retrieval becomes less reliable~(see fig.~\ref{real}), leading to MAE of $3.83^\circ$, higher than ideal case in previous work\cite{wang20253d}. In contrast, our proposed method achieves stable performance, delivering MAE of $1.56^\circ$, which also outperforms classical methods\cite{smith2018height,ba2020deep}~(see fig.\ref{real}).}

Moreover, we conduct a qualitative evaluation on a complex-shaped object (see fig.~\ref{horse}). For comparison, Fig.~\ref{horse}(a) presents the result obtained from the previous physics-based method~\cite{wang20253d}, which had to rely on \textbf{multi-shot} captures for this degree of object complexity.
As observed, the previous method exhibits noticeable noise and local inconsistencies, particularly in the cheek region of the horse, where the surface appears overly flat and lacks structural details. In contrast, our method, which operates in \textbf{single-shot} manner, produces a significantly more consistent normal field while preserving fine geometric structures~(see Fig.~\ref{horse}(b)). In addition, the inference time of the proposed method is $8 ms$, which is several orders of magnitude faster than the purely physics-based method.

\section{Conclusion and discussion:} The proposed method demonstrates generalization and robustness under single-shot acquisition while keeping the high accuracy. It achieves geometrically consistent full-field reconstruction even for surfaces with high spatial frequency structures and large curvature, where conventional single-shot methods fail to provide reliable measurement. Compared with previous analytical approaches, our method offers improved stability and faster inference in practical scenarios. 

Despite these encouraging results, several directions remain for further improvement. Acquiring reliable ground truth for real specular surfaces is inherently challenging, therefore, the network is trained on synthetic data generated from a digital twin. As a result, the performance on real measurements is slightly lower than that on synthetic data. This discrepancy can be attributed to real polarization camera factors that are not fully simulated by the current rendering model. Extending the training data to incorporate more realistic sensor characteristics would further enhance generalization. Future work may explore improved sensor-level modeling as well as hybrid training strategies that combine synthetic and real data.

In addition, the current framework is specifically designed for specular surfaces. Polarimetric 3D imaging is intrinsically influenced by material properties, and different materials may exhibit distinct polarization responses governed by different reflection models. Extending the proposed approach to handle a broader class of materials, including mixed or spatially varying reflectance, remains an important direction for future research.
We hope the proposed method will contribute to more practical and deployment-efficient 3D imaging solutions for complex specular surfaces in dynamic environments.
\\

\section*{Data Availability}
{The data that support the findings of this study are available from the corresponding author upon reasonable request..}

\section*{Conflict of Interest Statement}
The authors have no conflicts to disclose.

\bibliography{sample}

\end{document}